\documentclass[letterpaper, 10pt, conference]{./support/ieeeconf}
\usepackage{hyperref}
\usepackage{amsmath,amsfonts}
\usepackage[ruled,norelsize, linesnumbered]{algorithm2e}
\usepackage{array}
\usepackage{makecell}
\usepackage{subcaption}
\usepackage{textcomp}
\usepackage{stfloats}
\usepackage{url}
\usepackage{verbatim}
\usepackage{graphicx}
\usepackage{amssymb}
\usepackage{multirow}
\usepackage{booktabs}
\usepackage{lipsum}
\usepackage{float}
\usepackage{caption}
\usepackage[space]{cite}
\usepackage{colortbl}
\usepackage{pifont}
\newcommand{\cmark}{\ding{51}}
\newcommand{\xmark}{\ding{55}}

\usepackage[dvipsnames]{xcolor}

\IEEEoverridecommandlockouts                              
\title{\LARGE \bf
SIMPL: A Simple and Efficient Multi-agent Motion Prediction Baseline for Autonomous Driving
}

\author{Lu Zhang$^{1}$, Peiliang Li${^2}$, Sikang Liu${^2}$, and Shaojie Shen${^1}$
\thanks{$^{1}$L. Zhang and S. Shen are with the Department of Electronic and Computer Engineering, Hong Kong University of Science and Technology, Hong Kong (email: \{lzhangbz, eeshaojie\}@ust.hk). $^{2}$P. Li and S. Liu are with the DJI Technology Company, Ltd., Shenzhen, China (email: \{peiliang.li, sikang.liu\}@dji.com). Corresponding author: Lu Zhang.}
\thanks{This work was supported in part by the Hong Kong Ph.D. Fellowship Scheme, in part by the HKUST-DJI Joint Innovation Laboratory.}
}

\begin{document}
\maketitle
\thispagestyle{empty}
\pagestyle{empty}
\bstctlcite{IEEEexample:BSTcontrol}

\begin{abstract}
This paper presents a \underline{S}imple and eff\underline{I}cient \underline{M}otion \underline{P}rediction base\underline{L}ine (SIMPL) for autonomous vehicles. Unlike conventional agent-centric methods with high accuracy but repetitive computations and scene-centric methods with compromised accuracy and generalizability, SIMPL delivers real-time, accurate motion predictions for all relevant traffic participants. To achieve improvements in both accuracy and inference speed, we propose a compact and efficient global feature fusion module that performs directed message passing in a symmetric manner, enabling the network to forecast future motion for all road users in a single feed-forward pass and mitigating accuracy loss caused by viewpoint shifting. Additionally, we investigate the continuous trajectory parameterization using Bernstein basis polynomials in trajectory decoding, allowing evaluations of states and their higher-order derivatives at any desired time point, which is valuable for downstream planning tasks. As a strong baseline, SIMPL exhibits highly competitive performance on Argoverse 1 \& 2 motion forecasting benchmarks compared with other state-of-the-art methods. Furthermore, its lightweight design and low inference latency make SIMPL highly extensible and promising for real-world onboard deployment. We open-source the code at \url{https://github.com/HKUST-Aerial-Robotics/SIMPL}.
\end{abstract}
\setlength{\textfloatsep}{1pt}

\section{Introduction}
Motion forecasting for the surrounding traffic participants is essential in autonomous vehicles, especially for the downstream decision-making and planning modules, since accurate and timely intention and trajectory prediction will benefit both safety and riding comfort significantly.

For learning-based motion prediction, one of the most important topics is context representation. Early approaches typically represent the surrounding scene as a multi-channel bird's-eye-view image~\cite{cui2019multimodal, zhao2019multi, chai2019multipath, phan2020covernet}. In contrast, more recent research has increasingly embraced vectorized scene representations~\cite{gao2020vectornet, liang2020learning, liu2021multimodal, ngiam2021scene, zhou2022hivt, zhang2022trajectory, shi2022motion, feng2023macformer, gao2023dynamic}, in which the locations and geometries are annotated using point sets or polylines with geographic coordinates, leading to enhanced fidelity and expanded receptive fields. However, for both rasterized and vectorized representation, there exists a key question: how should we choose a suitable reference frame for all these elements? One straightforward way is to depict all instances within a shared coordinate system (scene-centric), such as one centered around the autonomous vehicle, and directly use the coordinates as the input features. This allows us to make predictions for multiple target agents in a single feed-forward pass~\cite{casas2020implicit, ngiam2021scene}. Yet, using global coordinates as the input, which often vary in a large span, will greatly intensify the inherent complexity of the task, resulting in degraded network performance and limited adaptability to novel scenarios. To achieve better accuracy and robustness, a common solution is normalizing the scene context w.r.t. the current state of the target agent~\cite{gao2020vectornet, liu2021multimodal, zhang2022trajectory, shi2022motion, feng2023macformer, gao2023dynamic} (agent-centric). It means the normalization process and feature encoding have to be executed repeatedly for each target agent, leading to better performance but at the cost of redundant computation. Hence, it is essential to explore an approach that can efficiently encode features for multiple targets while retaining robustness to changes of perspective.

\begin{figure}[t]
    \centering
    \includegraphics[width=0.48\textwidth]{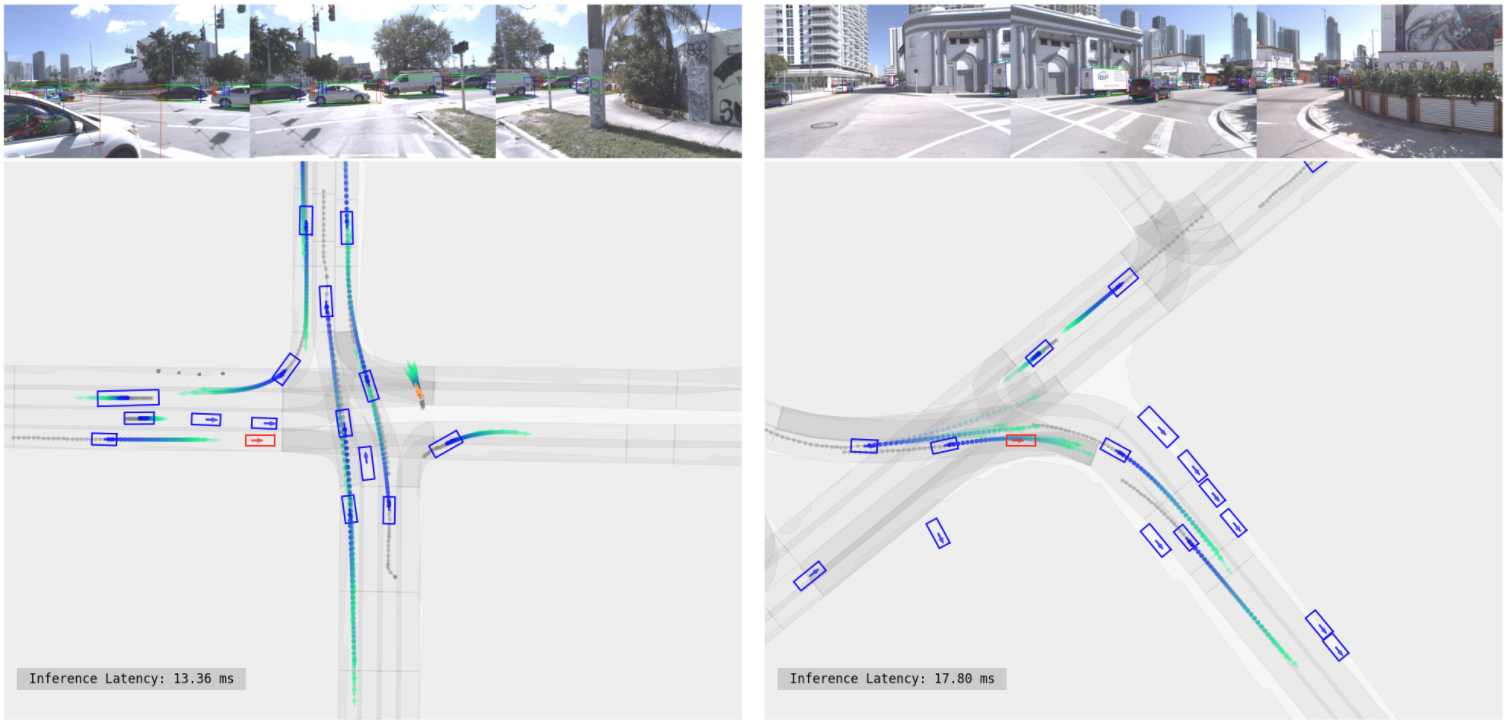}
    \caption{Illustration of multi-agent motion prediction in complex driving scenarios. Our method is able to generate reasonable hypotheses for all relevant agents simultaneously in a real-time fashion. The ego and other vehicles are shown in red and blue, respectively. Predicted trajectories are visualized using gradient color according to the timestamps. Please refer to the attached video for more examples.}\label{fig:cover}
\end{figure}

For the downstream modules of motion prediction, such as decision-making and motion planning, it is imperative to consider not only the future positions but also the headings, velocities, and other higher-order derivatives. For example, the predicted heading of surrounding vehicles plays a pivotal role in shaping the future spatiotemporal occupancy, a critical factor in ensuring safe and robust motion planning~\cite{werling2012optimal, ding2019safe}. In addition, the independent anticipation of higher-order quantities without adhering to physical constraints can introduce inconsistencies in prediction outcomes~\cite{buhet2020plop, jia2023towards}. For instance, it may generate positional displacement despite a zero velocity, leading to confusion in planning modules.

In this paper, we propose SIMPL (\textbf{S}imple and eff\textbf{I}cient \textbf{M}otion \textbf{P}rediction base\textbf{L}ine) for autonomous driving systems, addressing critical issues in multi-agent trajectory prediction for real-world onboard applications. Firstly, we introduce the instance-centric scene representation followed by a symmetric fusion Transformer (SFT), enabling efficient trajectory forecasting for all agents in a single feed-forward pass, while retaining accuracy and robustness brought by the viewpoint-invariant property. Compared with other recent works based on symmetric context fusion~\cite{jia2023hdgt, cui2023gorela, zhou2023query}, the proposed SFT is notably simpler, more lightweight, and easier to implement, making it suitable for onboard deployment.

Secondly, we introduce a novel parameterization method for the predicted trajectories based on the Bernstein basis polynomial (also known as the B\'{e}zier curve). This continuous representation ensures smoothness and enables effortless evaluation of exact states and their higher-order derivatives at any given time point. Our empirical studies indicate that learning to forecast the control points of B\'{e}zier curves is more effective and numerically stable compared to estimating the coefficients of monomial basis polynomials.

Lastly, the proposed components are well integrated into a simple and efficient model. We evaluate the proposed method on two large-scale motion forecasting datasets~\cite{chang2019argoverse, wilson2023argoverse}, and the experimental results show that SIMPL is highly competitive when compared with other state-of-the-art methods despite its streamlined design. More importantly, SIMPL achieves efficient multi-agent trajectory prediction with fewer learnable parameters and lower inference latency without sacrificing quantitative performance, which is promising for real-world onboard deployment. We also highlight that SIMPL gains excellent extensibility as a strong baseline. The succinct architecture facilitates straightforward integration with recent advances in motion forecasting, offering opportunities for further enhancements in overall performance.

\section{Related Work}\label{sec:related_work}
\subsection{Context Encoding and Fusion}
Driving context can be broadly categorized into two main types: the historical trajectories of surrounding agents and static map information. Trajectories, as time series data, are typically encoded by temporal networks~\cite{alahi2016social, vemula2018social}. As for the map features, early works commonly represent it as multi-channel bird's-eye-view images with different semantic elements rendered in distinct channels, then utilize convolutional neural networks (CNNs) to perform feature fusion~\cite{cui2019multimodal, zhao2019multi, chai2019multipath, phan2020covernet}. However, the rasterization inevitably introduces information loss and leads to limited receptive fields. To address these issues, vectorized-based methods are proposed~\cite{gao2020vectornet, liang2020learning} and become increasingly prevalent~\cite{liu2021multimodal, ngiam2021scene, zhou2022hivt, zhang2022trajectory, shi2022motion, feng2023macformer, gao2023dynamic}. In such methods, map elements are represented as polylines~\cite{gao2020vectornet, liu2021multimodal, zhou2022hivt, shi2022motion} and sparse graphs~\cite{liang2020learning, zhang2022trajectory, gao2023dynamic}, preserving spatial information using raw coordinates. These features are further processed via graph neural networks~\cite{kipf2016semi, velivckovic2017graph} or Transformers~\cite{vaswani2017attention}, yielding higher fidelity and better efficiency.

\subsection{Symmetric Scene Modeling}
Both scene-centric~\cite{casas2020implicit, ngiam2021scene} and agent-centric~\cite{gao2020vectornet, liu2021multimodal, zhang2022trajectory, shi2022motion, feng2023macformer, gao2023dynamic} representations have their limitations, necessitating a trade-off between accuracy and computational overhead. Recently, several approaches~\cite{zhou2022hivt, jia2023hdgt, cui2023gorela, zhou2023query} have emerged to address this issue by introducing symmetric modeling into the feature fusion process. HiVT~\cite{zhou2022hivt} normalizes the local context for each agent and explicitly incorporates relative poses in both local and global feature fusion, making the method viewpoint-invariant. HDGT~\cite{jia2023hdgt} and GoRela~\cite{cui2023gorela} introduce the pairwise relative positional encoding in the message passing of the heterogeneous graphs. Taking one step further, QCNet~\cite{zhou2023query} extends the viewpoint-invariant property to the spatial-temporal domain by incorporating the time dimension into the relative positional encoding, enabling support for streaming processing. Compared with these approaches, our work also adopts a similar idea but proposes a compact symmetric feature fusion module, which is distinctly simpler, more lightweight, and easier to implement.

\subsection{Trajectory Representation}
Predicted trajectories are commonly represented as sequences of discrete states, such as positional coordinates~\cite{gao2020vectornet, liang2020learning} or mixtures of probability distributions~\cite{chai2019multipath, zhou2022hivt}. Since there is no explicit constraint between discrete states, this always leads to jagged, kinematically infeasible trajectories. An alternative method predicts control signals and integrates them recurrently into trajectories according to kinematic models~\cite{cui2020deep, vara2022multipath++}. However, this recurrent formulation tends to be less efficient and can be more susceptible to perception errors. Continuous trajectory parameterization, such as B\'{e}zier curves, is widely used in trajectory planning for mobile robots. One can efficiently generate a smooth and continuous optimal trajectory by manipulating the control points while considering certain objectives and constraints~\cite{gao2018online,deolasee2023spatio}. In this paper, we leverage B\'{e}zier curves as the output form, which ensures single-step decoding without recurrent unrolling while maintaining superior numerical stability compared to the monomial basis polynomials~\cite{buhet2020plop}.

\section{Methodology}\label{sec:method}
\subsection{Problem Formulation}
The trajectory prediction task involves generating potential future trajectories for the target agents based on the observed motion history of moving objects and the surrounding map information. Specifically, in a driving scenario with $N_a$ moving agents (including AV), we use $\mathcal{M}$ to represent the map information and use $\mathbf{X}=\{\mathbf{x}_0,\dots,\mathbf{x}_{N_a}\}$ to collectively denote the observed trajectories of all agents. Here, each $\mathbf{x}_i=\{x_{i,-H+1},\dots,x_{i,0}\}$ represents the historical trajectory of the $i$-th agent over the past $H$ time steps. Without loss of generality, the multi-agent motion predictor generates potential future trajectories for all $N_a$ agents in the scene, represented as $\mathbf{Y}=\{\mathbf{y}_0,\dots,\mathbf{y}_{N_a}\}$. For each individual agent $i$, $K$ possible future trajectories and their corresponding probability scores are predicted to capture the inherent multimodal distribution. The multimodal trajectories are represented as $\mathbf{y}_i=\{\mathbf{y}^{1}_{i},\dots,\mathbf{y}^{K}_{i}\}$, where each $\mathbf{y}^{k}_i=\{y^{k}_{i,1},\dots,y^{k}_{i,T}\}$ is the $k$-th predicted trajectory of the $i$-th agent over the prediction horizon $T$, while the probability score list is represented as $\mathbf{\alpha}_i=\{\alpha^1_{i},\dots,\alpha^K_{i}\}$. Hence, the multimodal trajectory prediction for agent $i$ can be regarded as estimating a mixture distribution
\begin{equation}
P(\mathbf{y}_i|\mathbf{X},\mathcal{M})=\sum^{K}_{k=1}\alpha^k_iP(\mathbf{y}^k_i|\mathbf{X},\mathcal{M}).\nonumber
\end{equation}
Note that we primarily focus on marginal motion prediction in this paper, but our approach can be smoothly extended to joint motion prediction tasks by involving scene-level loss functions~\cite{ngiam2021scene, girgis2021latent}. We leave it as an important future work.

\begin{figure*}[t]
    \centering
    \includegraphics[width=0.85\textwidth]{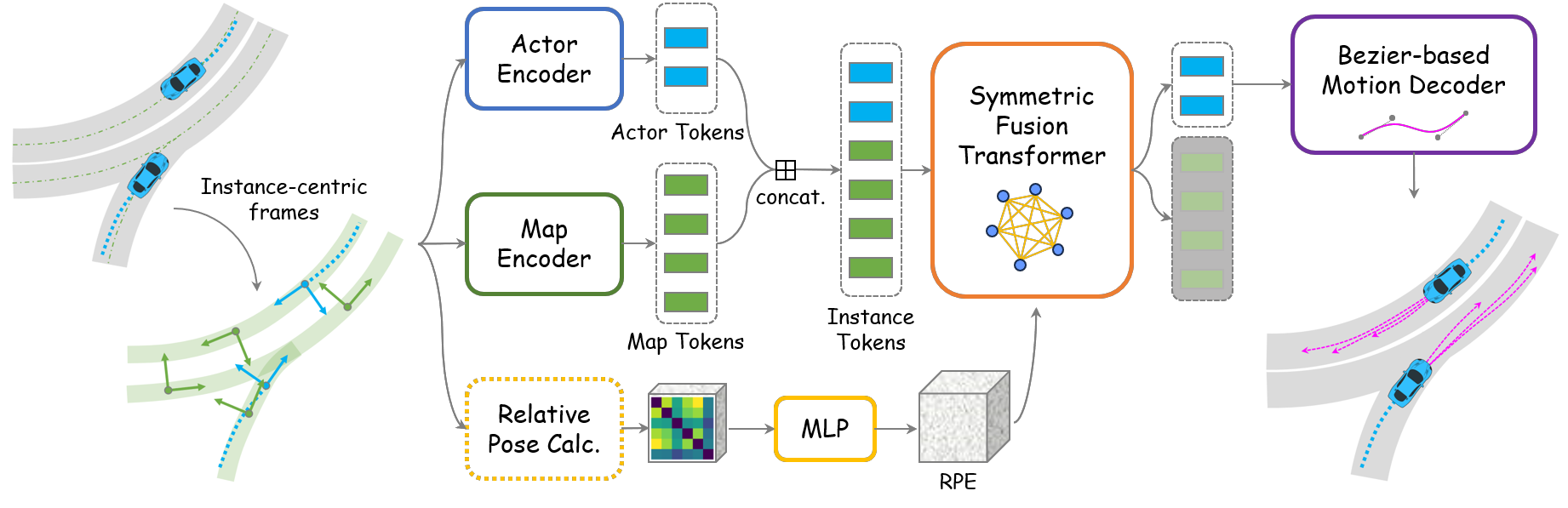}
    \caption{Illustration of SIMPL. We utilize the simplest possible network architecture to demonstrate its effectiveness. The local features of semantic instances are processed by simple encoders, while the inter-instance features are preserved in the relative positional embeddings. Multimodal trajectory prediction results are generated by the motion decoder after the proposed symmetric feature Transformer.}\label{fig:framework}
\vspace{-0.6cm}
\end{figure*}

\begin{figure}[t]
	\centering
	\includegraphics[width=0.42\textwidth]{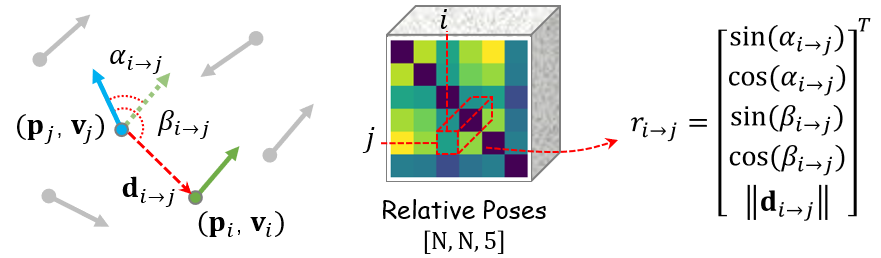}
	\caption{Illustration of the relative pose calculation. A typical scene is depicted on the left, and we leave out the $y$-axis of the anchor poses for conciseness. The relative pose between instance $i$ and $j$ can be described by the heading difference $\alpha_{i\rightarrow j}$, relative azimuth $\beta_{i\rightarrow j}$, and positional distance $\lVert\mathbf{d}_{i\rightarrow j}\rVert$. The \textit{all-to-all} relative poses are calculated and formulated as a 3D array.}\label{fig:rpe}
\end{figure}

\subsection{Framework Overview}
An overview of the proposed SIMPL framework is shown in Fig.~\ref{fig:framework}. Firstly, we adopt the vectorized scene representation. For each semantic instance, such as trajectories and lane segments, we construct a local reference frame to decouple inherent features and relative information between instances. Then, the actor and map features are extracted by simple encoders, while the relative poses of instances are calculated in pairs and further encoded by a multilayer perceptron (MLP) to get the relative positional embedding (RPE). Instance tokens and RPE are then sent into the proposed symmetric fusion Transformer (SFT), a compact and succinct fusion module that symmetrically updates features. Finally, the B\'{e}zier curve parameterized trajectories are predicted by a simple decoder for all target agents simultaneously.

\subsection{Instance-centric Scene Representation}
Apart from scene-centric representation, a scenario can be represented by vectorized features under the local frames of the instances, along with the relative poses between them. A local reference frame is established for each semantic element to normalize its spatial attributes, which we term as ``instance-centric". Without loss of generality, we locate the reference frame at the current observed state for agents' historical trajectories. Regarding static map elements, such as lane segments, we use the centroid of the polyline as the anchor point and employ the displacement vector between endpoints as the heading angle. Intuitively, the local coordinate frames can be regarded as the ``anchor pose" of the instances, therefore, the relative spatial information can be easily calculated pair-by-pair.

Specifically, the anchor pose under a global coordinate frame for element $i$ can be represented using its position $\mathbf{p}_i\in\mathbb{R}^2$ and heading vector $\mathbf{v}_i\in\mathbb{R}^2$. Following~\cite{cui2023gorela}, we describe the relative pose between element $i$ and element $j$ using three quantities: heading difference $\alpha_{i\rightarrow j}$, relative azimuth $\beta_{i\rightarrow j}$, and distance $\lVert\mathbf{d}_{i\rightarrow j}\rVert$. To enhance numerical stability, angles are represented using their sine and cosine values. We denote the heading difference $\alpha_{i\rightarrow j}$ as
\begin{equation}
    \sin(\alpha_{i\rightarrow j}) = \frac{\mathbf{v}_i \times \mathbf{v}_j}{\lVert\mathbf{v}_i\rVert\lVert\mathbf{v}_j\rVert}, \quad \cos(\alpha_{i\rightarrow j}) = \frac{\mathbf{v}_i \cdot \mathbf{v}_j}{\lVert\mathbf{v}_i\rVert\lVert\mathbf{v}_j\rVert}, \nonumber
\end{equation}
and the relative azimuth $\beta_{i\rightarrow j}$ (the angle between displacement vector $\mathbf{d}_{i\rightarrow j}=\mathbf{p}_i-\mathbf{p}_j$ and heading vector $\mathbf{v}_j$) as
\begin{equation}
    \sin(\beta_{i\rightarrow j}) = \frac{\mathbf{d}_{i\rightarrow j} \times \mathbf{v}_j}{\lVert\mathbf{d}_{i\rightarrow j}\rVert\lVert\mathbf{v}_j\rVert}, \quad \cos(\beta_{i\rightarrow j}) = \frac{\mathbf{d}_{i\rightarrow j} \cdot \mathbf{v}_j}{\lVert\mathbf{d}_{i\rightarrow j}\rVert\lVert\mathbf{v}_j\rVert}. \nonumber
\end{equation}
For simplicity, we omit the additional positional encoding process for the distance value used in~\cite{cui2023gorela}, making the relative spatial information a 5-dimensional vector $r_{i\rightarrow j} = \left[ \sin(\alpha_{i\rightarrow j}), \cos(\alpha_{i\rightarrow j}), \sin(\beta_{i\rightarrow j}), \cos(\beta_{i\rightarrow j}), \lVert\mathbf{d}_{i\rightarrow j}\rVert \right]$. We can conveniently calculate the \textit{all-to-all} relative spatial information by leveraging the broadcasting mechanism of PyTorch or NumPy. Consequently, given a scene contains $N=N_a+N_m$ semantic elements, the resulting relative positional info is an array with the shape of $\left[N, N, 5\right]$, while $r_{i\rightarrow j}$ locates at the $j$-th row and $i$-th column. An illustration of the relative pose calculation is shown in Fig~\ref{fig:rpe}.

\subsection{Context Feature Encoding}
After obtaining the instance-centric representation and relative positional encoding for instances, we utilize corresponding encoders (also serving as ``tokenizers") to convert them into feature vectors. To keep SIMPL simple, we use the 1D CNN-based network~\cite{liang2020learning} for handling historical trajectories and employ a PointNet-based encoder~\cite{qi2017pointnet, gao2020vectornet} for extracting static map features. Without loss of generality, we let all latent features have $D$ channels. Therefore, the resulting actor and map tokens have the shapes of $\left[N_a, D\right]$ and $\left[N_m, D\right]$, where $N_a$ is the number of actors and $N_m$ is the number of map elements. For the detailed implementation, we refer readers to~\cite{liang2020learning, gao2020vectornet}. Moreover, the relative pose encoding is further encoded by an MLP, yielding the relative positional embedding (RPE) with the shape of $\left[N, N, D\right]$.

\subsection{Symmetric Fusion Transformer}
\begin{figure}[!t]
	\centering
	\includegraphics[width=0.48\textwidth]{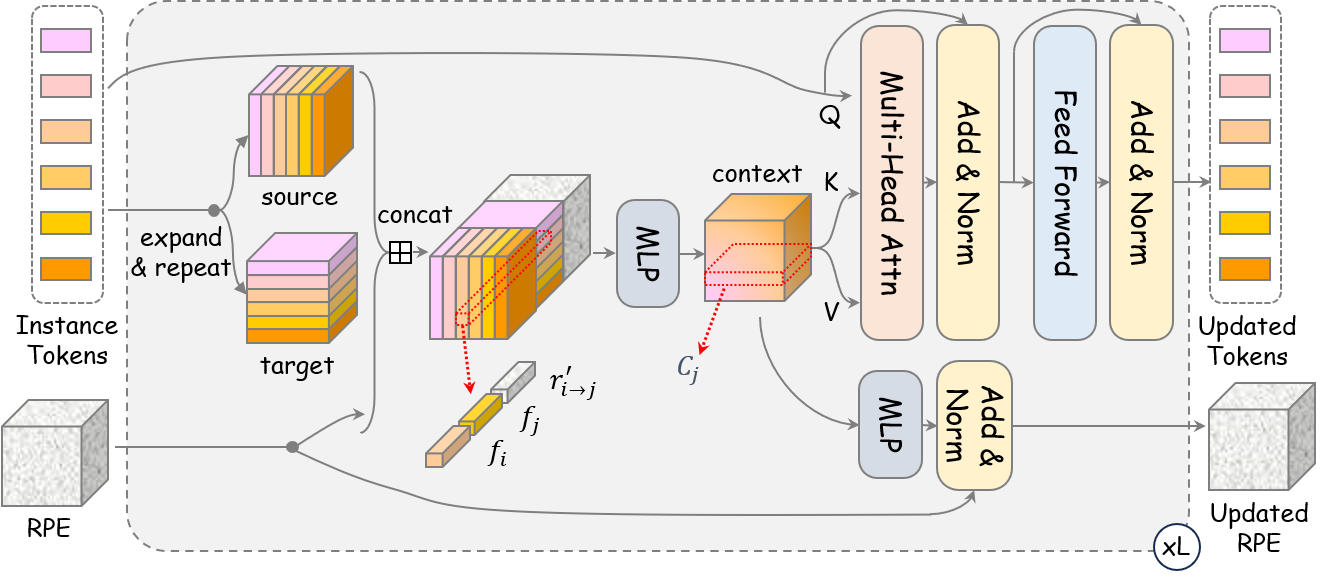}
	\caption{Illustration of the proposed symmetric fusion Transformer (SFT) with $L$ layers. Instance tokens and RPE are recurrently updated in each SFT layer.}\label{fig:sft}
\end{figure}

Once the instance tokens and the corresponding RPE are obtained, we employ the proposed symmetric fusion Transformer (SFT) to update the instance tokens in a viewpoint-invariant manner. Fig.~\ref{fig:sft} shows the overall structure of the proposed SFT, which comprises multiple stacked SFT layers, akin to the standard Transformer~\cite{vaswani2017attention}. In essence, we can view the driving scene as a complete digraph with self-loops, in which the input instance-centric features serve as nodes and the RPE depicts the edge information. During the update process, the node features are influenced solely by the graph edges associated with the target node, ensuring that the feature fusion remains viewpoint-invariant.

In the \textit{microscopic} view, we denote the token of $i$-th and $j$-th instances as $f_i$ and $f_j$, respectively. And the RPE vector associated with the edge from $f_i$ to $f_j$ is designated as $r'_{i\rightarrow j}$. The tuple $\left(f_i, f_j, r'_{i\rightarrow j}\right)$ encompasses all information intended to be transmitted from node $i$ to node $j$, therefore, we can employ a simple MLP to encode these features and get the $i$-th context vector for node $j$
\begin{equation}
    c_{i\rightarrow j} = \phi\left( f_i \boxplus f_j \boxplus r'_{i\rightarrow j} \right), \nonumber
\end{equation}
where $\boxplus$ denotes the concatenation operator, and $\phi:\mathbb{R}^{3D}\rightarrow\mathbb{R}^D$ denotes the MLP, which consists of a linear layer, layer normalization, and ReLU activation. We then perform cross-attention on the target node and its context, 
\begin{equation}
    f'_j = \texttt{MHA}(\texttt{Query}:f_j, \texttt{Key}:C_j, \texttt{Value}:C_j), \nonumber
\end{equation}
where $\texttt{MHA}(\cdot,\cdot,\cdot)$ is the standard multi-head attention function, and $C_j=\{c_{i\rightarrow j}\}_{i\in\{1,\dots,N\}}$ is the set of context vectors of token $j$. Note that $C_j$ contains $c_{j\rightarrow j}$ as well, indicating the presence of a self-loop for each node. Same as the standard Transformer, a point-wise feed-forward layer is integrated after the attention mechanism. Additionally, in each layer, $r'_{i\rightarrow j}$ is updated by re-encoding the context vector using another MLP and subsequently added to the input RPE through the residual connection.

In practice, we provide a more efficient implementation of the aforementioned feature fusion in a vectorized way (see Fig.~\ref{fig:sft}). Firstly, given the input instance tokens $F\in\mathbb{R}^{N\times D}$, we expand it along different dimensions and replicate it $N$ times to build the source and target node arrays, which both exhibit the shape of $[N, N, D]$. After concatenation of the source array, target array, and the corresponding RPE, the array of tuple $\left(f_i, f_j, r'_{i\rightarrow j}\right)$ is obtained, and we apply $\phi$ to get the context array $C\in\mathbb{R}^{N\times N\times D}$. Note that the $j$-th row of $C$ is exactly $C_j$, representing the collection of context features centered around token $j$. Hence, we employ $C$ for the key and value, while the expanded $F\in\mathbb{R}^{N\times 1\times D}$ serves as the query. The standard multi-head attention module then passes messages from context features to instance tokens. The rest of the SFT layer also enjoys the vectorized implementation, but we won't delve into the details due to its simplicity. It's worth noting that our proposed SFT layer shares similarities with recent ``query-centric" methods~\cite{zhou2023query, shi2023mtr++}, but we incorporate global attention and RPE updates, resulting in a more compact design. For the detailed implementation, please refer to the released code.

\subsection{Multimodal Continuous Trajectory Decoder}
After the symmetric global feature fusion, the updated actor tokens are gathered and sent to a multimodal motion decoder to generate predictions for all agents. Here, we forecast $K$ possible futures, and for each mode, a simple MLP is applied, which has a regression head for trajectories and a classification head followed by a softmax function for their corresponding probability scores. 

As for the trajectory regression head, in contrast to previous approaches that directly predict the positions of future trajectories, we choose to use the continuous parameterized representation. Parameterized curves (e.g., polynomials) bring a continuous representation, which allows for obtaining smooth motion and exact high-order derivatives at any time point. However, according to the previous studies~\cite{vara2022multipath++}, the monomial basis polynomial representation hurts performance significantly. We blame the degeneration on the numerical imbalance of the predicted coefficients (see Sec.~\ref{sec:abl_param} for details), making the regression a hard task.

\begin{figure}[t]
	\centering
	\includegraphics[width=0.48\textwidth]{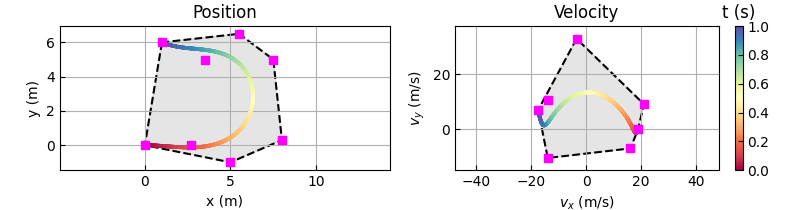}
	\caption{A 2D septic B\'{e}zier curve (left). Pink dots are control points while grey polygons are corresponding convex hulls. When the time duration of the trajectory is 1 second, the $1$st-order derivative will exactly be the velocity profile (right), which is also a B\'{e}zier curve due to the hodograph property.}\label{fig:bezier}
\end{figure}

To leverage the advantage of parametric trajectory while avoiding performance degeneration, we introduce the Bernstein basis polynomial (i.e., B\'{e}zier curve), of which coefficients are control points with concrete spatial meaning, resulting in better convergence. Specifically, an degree $n$ B\'{e}zier curve is written as
\begin{equation}
    f(t) = \sum_{i=0}^nb_{n}^i(t)p_i = \sum_{i=0}^n\binom{n}{i}t^i(1-t)^{n-i}p_{i},~t\in[0,1], \nonumber
\end{equation}
where $b_{n}^i(t)$ is the $i$-th order Bernstein basis, $\binom{n}{i}$ is the binomial coefficient, $t$ is the variable of the parametric curve, and $p_i$ is the control point. Note that for an $n$-th order B\'{e}zier curve, there are $n+1$ control points in total, and the first and last control points are always the endpoints of the curve (see Fig.~\ref{fig:bezier}). As B\'{e}zier curves are defined on $t\in[0,1]$, we normalize the actual time $\tau\in[0,\tau_{max}]$, so the parametric curve can be written as $f(t) = \sum_{i=0}^nb_{n}^i(\frac{\tau}{\tau_{max}})p_i$. Moreover, due to the hodograph property, the derivative of an $n$th-order B\'{e}zier curve is still a B\'{e}zier curve, with control points defined by $p^{(1)}_{i} = n(p_{i+1} - p_i)$, namely, the velocity profile of the trajectory can be calculated by
\begin{equation}
    f'(t) = \frac{n}{\tau_{max}}\sum_{i=0}^{n-1}b_{n-1}^i(\frac{\tau}{\tau_{max}})(p_{i+1} - p_i),~\tau\in[0,\tau_{max}]. \nonumber
\end{equation}

In practice, we use a simple MLP as the regression head that performs the mapping from the fused actor features to the control points. Then, the positional coordinates of each predicted trajectories $\mathbf{Y}_{pos}\in\mathbb{R}^{T\times2}$ can be simply calculated by multiplying the constant basis matrix $\mathbf{B}\in\mathbb{R}^{T\times (n+1)}$ and the corresponding predicted 2D control points $\mathbf{P}\in\mathbb{R}^{(n+1)\times 2}$ (independent for $x$-axis and $y$-axis),
\begin{align*}
    \mathbf{Y}_{pos} &= \mathbf{B} \times \mathbf{P} \\
                     &=
    \begin{bmatrix}
    b_{n}^0(t_1) & b_{n}^1(t_1) & ... & b_{n}^n(t_1) \\
    \vdots       & \vdots       &     & \vdots \\
    b_{n}^0(t_{T}) & b_{n}^1(t_{T}) & ... & b_{n}^n(t_{T}) \\
    \end{bmatrix}
    \begin{bmatrix}
    p^x_0   & p^y_0  \\
    \vdots  & \vdots \\
    p^x_n   & p^y_n  \\
    \end{bmatrix},
\end{align*}
where ${T}$ is the number of sampled timestamps required for the predicted trajectories, and $t_i=\frac{\tau_i}{\tau_{max}}$ is the normalized time point. We also point out that the velocity and other higher-order derivatives of the predicted trajectories can be retrieved by the similar procedures mentioned above and we omit it for conciseness. For agents with non-holonomic constraints, like vehicles and cyclists, the yaw angle aligns with the tangent vector of the trajectory, and we can derive the heading angle for each state from the velocity estimation. Finally, the predicted trajectories are further transformed back to the global coordinates according to the corresponding anchor poses of the actors.

\subsection{Training}\label{sec:training}
The proposed SIMPL is trained in an end-to-end manner. The overall loss function is the weighted sum of the regression loss and classification loss
\begin{equation}
    \mathcal{L} = \omega \mathcal{L}_{reg} + (1-\omega) \mathcal{L}_{cls}, \nonumber
\end{equation}
where $\omega\in[0,1]$ is the weight to balance these components, and we set $\omega=0.8$ to address the importance of the regression task. Following~\cite{liang2020learning}, we use the winner-takes-all (WTA) strategy for handling the multimodality. For each agent, we find the best-predicted trajectory $k^*$ among the $K$ hypotheses by picking the one with minimum final displacement error. Regarding the classification task, we use the max-margin loss to distinguish the positive mode from others similar to~\cite{liang2020learning}. For the trajectory regression task, in addition to positional coordinate regression, we introduce an optional yaw angle loss to provide auxiliary supervision, resulting in
\begin{equation}
    \mathcal{L}_{reg} = \texttt{PosLoss}(\bar{Y}_{pos}, Y^{k^*}_{pos}) + \texttt{YawLoss}(\bar{Y}_{yaw}, Y^{k^*}_{yaw}), \nonumber
\end{equation}
where $\bar{Y}_{(\cdot)}$ denotes the ground truth (GT) states, and $Y^{k^*}_{(\cdot)}$ is the predicted position and yaw angle of the winner mode. We employ the smooth L1 loss as the position regression loss, and we designate the yaw regression loss as
\begin{equation}
    \texttt{YawLoss}(\bar{Y}_{yaw}, Y^{k^*}_{yaw}) = [1 - \texttt{CosSim}(\bar{Y}_{yaw}, Y^{k^*}_{yaw})]/2, \nonumber
\end{equation}
where $\texttt{CosSim}(\cdot,\cdot)$ is the cosine similarity measurement, which yields a value of $1$ for two aligned yaw vectors and a value of $-1$ for two opposite yaw vectors. Incorporating yaw angle loss implicitly strengthens the consistency between consecutive states, making predicted trajectories with higher smoothness and kinematic feasibility, and resulting in more realistic trajectories, especially for low-speed agents.

\section{Experimental Results}\label{sec:exp_results}
\subsection{Experiment Setup}
\subsubsection{Dataset}
We evaluate the proposed method on both Argoverse 1~\cite{chang2019argoverse} and Argoverse 2~\cite{wilson2023argoverse} motion forecasting datasets. Argoverse 1 contains 205942, 39472, and 78143 sequences for training, validation, and testing, respectively. Each sequence is sampled at 10 Hz, and the task involves predicting future 3-second trajectories based on 2 seconds of historical observations (i.e., $H=20$, $T=30$). In the case of Argoverse 2, it comprises 200000, 25000, and 25000 sequences for training, validation, and testing. The sequences are also sampled at 10 Hz, and the given history is 5 seconds while the future motion is 6 seconds (i.e., $H=50$, $T=60$). Both Argoverse 1 and Argoverse 2 provide HD maps.

\subsubsection{Metrics}
We mainly follow the standard metrics that are commonly used in multimodal trajectory prediction, including minimum average displacement error (minADE$_k$), minimum final displacement error (minFDE$_k$), miss rate (MR$_k$), and brier-minFDE$_k$. All these metrics evaluate the best-predicted trajectory for a single target agent among the $K$ hypotheses against the ground truth. The minADE$_k$ is the average Euclidean distance between the predicted trajectory and GT, while minFDE$_k$ only considers the error at endpoints. The MR is the percentage of sequences where the obtained minFDE$_k$ is greater than 2 meters. Brier-minFDE$_k$ adds an additional brier score $(1-p)^2$ to minFDE$_k$, where $p$ denotes the probability of the best-predicted trajectory. We refer interested readers to~\cite{chang2019argoverse, wilson2023argoverse} for detailed definitions. 

\subsubsection{Implementation details}
We set the dimension $D=128$ for all latent vectors, and stack 4 SFT layers and 8 attention heads for the symmetric global feature fusion. For the multimodal decoder, we set the number of modes $K=6$ following the common setups. The degree of B\'ezier curve $n$ is configured as 5 for Argoverse 1 and 7 for Argoverse 2 due to the different prediction horizons. SIMPL is trained in an end-to-end manner using a batch size of 128 for 50 epochs on a server with 8 Nvidia RTX 3090 GPUs. We employ the Adam optimizer and set the learning rate to 1e-3 in the beginning and gradually decrease it to 1e-4 after 40 epochs.

\subsection{Results}
\subsubsection{Comparison with the state-of-the-art}
We compare SIMPL with other state-of-the-art methods on two large-scale motion forecasting benchmarks. Tab.~\ref{tab:av1_test} shows the quantitative results of the Argoverse 1 test split. The upper part presents the single-model results, while the lower part shows the performance of methods with ensemble techniques. With such a simple design, SIMPL achieves highly competitive results among all listed methods. LaneGCN~\cite{liang2020learning}, mmTransformer~\cite{liu2021multimodal}, and MacFormer~\cite{feng2023macformer} utilize the agent-centric representation, which hinders efficient online inference. Scene Transformer~\cite{ngiam2021scene} adopts a scene-centric representation, enabling single-pass multi-agent motion prediction. However, it exhibits a larger model size and inferior performance, indicating it is data-hungry and less generalizable. HiVT~\cite{zhou2022hivt} explicitly considers relative poses during feature fusion for robustness against viewpoint shifting, whereas SIMPL yields a simpler and lighter design while achieving better performance. We also report the evaluation results with ensembling for a fair comparison. After ensembling of 8 models based on k-means clustering, SIMPL outperforms strong baselines like MultiPath++~\cite{vara2022multipath++} and is competitive to state-of-the-art methods such as Wayformer~\cite{nayakanti2023wayformer} with much fewer parameters. The evaluation results of the Argoverse 2 motion forecasting benchmark are shown in Tab.~\ref{tab:av2_test}. We compare SIMPL with other state-of-the-art methods that employ symmetric scene modeling techniques. Characterized by its minimalist architecture and remarkably compact model size, SIMPL attains competitive trajectory prediction results and is promising for further extensions and applications.

\begin{table}[tb]
	\centering
	\caption{Results on the test split of Argoverse 1 motion forecasting dataset. The upper and lower groups are the results of single model and ensemble methods. The best result is in \textbf{bold} while the second best result is \underline{underlined}. b-minFDE$_{6}$ is the official ranking metric. $\sharp$ denotes the model size is from the non-official implementation\label{tab:av1_test}}
	\setlength{\tabcolsep}{1.2mm}
	\begin{tabular}{l|ccc >{\columncolor[gray]{0.9}}c|c}
	\toprule
	Method                              & minADE$_{6}$      & minFDE$_{6}$      & MR$_{6}$         & b-minFDE$_{6}$    & \#Param \\
	\midrule
    LaneGCN~\cite{liang2020learning}	& 0.870             & 1.362             & 16.2             & 2.053             & 3.7M  \\
    mmTrans~\cite{liu2021multimodal}    & 0.844             & 1.338             & 15.4             & 2.033             & 2.6M  \\
    SceneTrans~\cite{ngiam2021scene}    & 0.803             & 1.232             & 12.6             & 1.887             & 15.3M \\
    HiVT~\cite{zhou2022hivt}            & \textbf{0.774}    & \textbf{1.169}    & 12.7             & 1.842             & 2.5M  \\
    MacFormer~\cite{feng2023macformer}  & 0.819             & 1.216             & \textbf{12.1}    & \underline{1.827} & 2.4M  \\
    SIMPL (w/o ens)                     & \underline{0.793} & \underline{1.179} & \underline{12.3} & \textbf{1.809}    & \textbf{1.8M}     \\
    \midrule
    MultiPath++~\cite{vara2022multipath++}  & 0.790             & 1.214             & 13.2             & 1.793             & 21.1M$^\sharp$   \\
    MacFormer~\cite{feng2023macformer}      & 0.812             & 1.214             & 12.7             & 1.767             & 2.4M \\
    HeteroGCN~\cite{gao2023dynamic}         & 0.789             & 1.160             & \underline{11.7} & 1.751             & -    \\
    Wayformer~\cite{nayakanti2023wayformer} & \textbf{0.768}    & \underline{1.162} & 11.9             & \textbf{1.741}    & 11.2M$^\sharp$   \\
    SIMPL (w/ ens)                          & \underline{0.769} & \textbf{1.154}    & \textbf{11.6}    & \underline{1.746} & \textbf{1.8M}    \\
  \bottomrule
\end{tabular}
\end{table}

\begin{table}[tb]
	\centering
	\caption{Results on the Argoverse 2 test split for methods based on symmetric scene modeling. The results are from single models (w/o ensemble). The best and the second-best results are in \textbf{bold} and \underline{underlined}, respectively. b-minFDE$_{6}$ is the official ranking metric.\label{tab:av2_test}}
	\setlength{\tabcolsep}{1.2mm}
	\begin{tabular}{l|ccc >{\columncolor[gray]{0.9}}c|c}
	\toprule
	Method                      & minADE$_{6}$     & minFDE$_{6}$     & MR$_{6}$         & b-minFDE$_{6}$      & \#Param        \\
	\midrule
    HDGT~\cite{jia2023hdgt}	    & 0.84             & 1.60             & 21.0             & 2.24                &  12.1M         \\
    GoRela~\cite{cui2023gorela}	& 0.76             & 1.48             & 22.0             & \underline{2.01}    &  -             \\
    QCNet~\cite{zhou2023query}	& \textbf{0.65}    & \textbf{1.29}    & \textbf{16.0}    & \textbf{1.91}       &  7.3M          \\
    SIMPL (w/o ens)             & \underline{0.72} & \underline{1.43} & \underline{19.2} & 2.05                &  \textbf{1.9M} \\
  \bottomrule
\end{tabular}
\end{table}

\subsubsection{Inference latency}
\begin{figure}[t]
	\centering
	\includegraphics[width=0.48\textwidth]{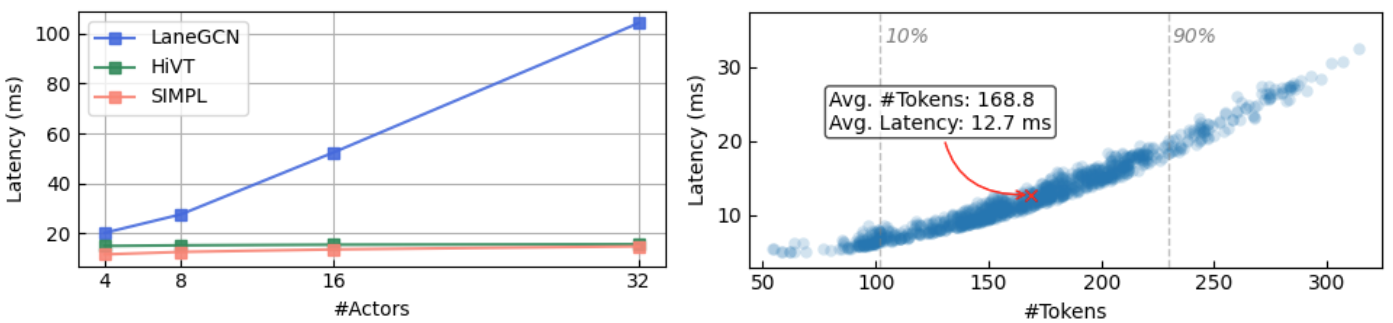}
	\caption{Evaluation results of the inference latency on Argoverse 1 dataset. Left: Average inference latency of different methods w.r.t. the number of target agents. Both HiVT and SIMPL attain real-time performance for multi-agent motion prediction, while agent-centric approaches are hard to scale up. Right: Relation between runtime and number of total instance tokens in the scenes. Each point represents a driving scene and all of them can be processed in real-time.}\label{fig:qual_latency}
\end{figure}

The evaluation results of inference latency are shown in Fig.~\ref{fig:qual_latency}. All experiments are conducted on an RTX 3060Ti GPU with the original PyTorch implementation. Firstly, we compare the computational efficiency with LaneGCN~\cite{liang2020learning} and HiVT~\cite{zhou2022hivt}. As the agent-centric baseline, LaneGCN normalizes the scene w.r.t. each target's state and organizes contexts into a batched form. Both HiVT and our SIMPL employ shared context encoding and perform multi-agent prediction in one forward pass, meaning that the batch size can be set as 1. Benefiting from the compact design, SIMPL achieves real-time performance and slightly better inference speed than HiVT. Besides, methods based on symmetric scene modeling enable much more efficient multi-agent prediction than conventional agent-centric counterparts. The right part of Fig.~\ref{fig:qual_latency} shows the latency distribution on 1,000 scenes randomly sampled from the full validation set. Without any acceleration techniques like quantization, SIMPL attains high inference speed and is promising for real-world onboard deployment after further optimization.

\subsubsection{Qualitative Results}
\begin{figure*}[t]
	\centering
	\includegraphics[width=0.98\textwidth]{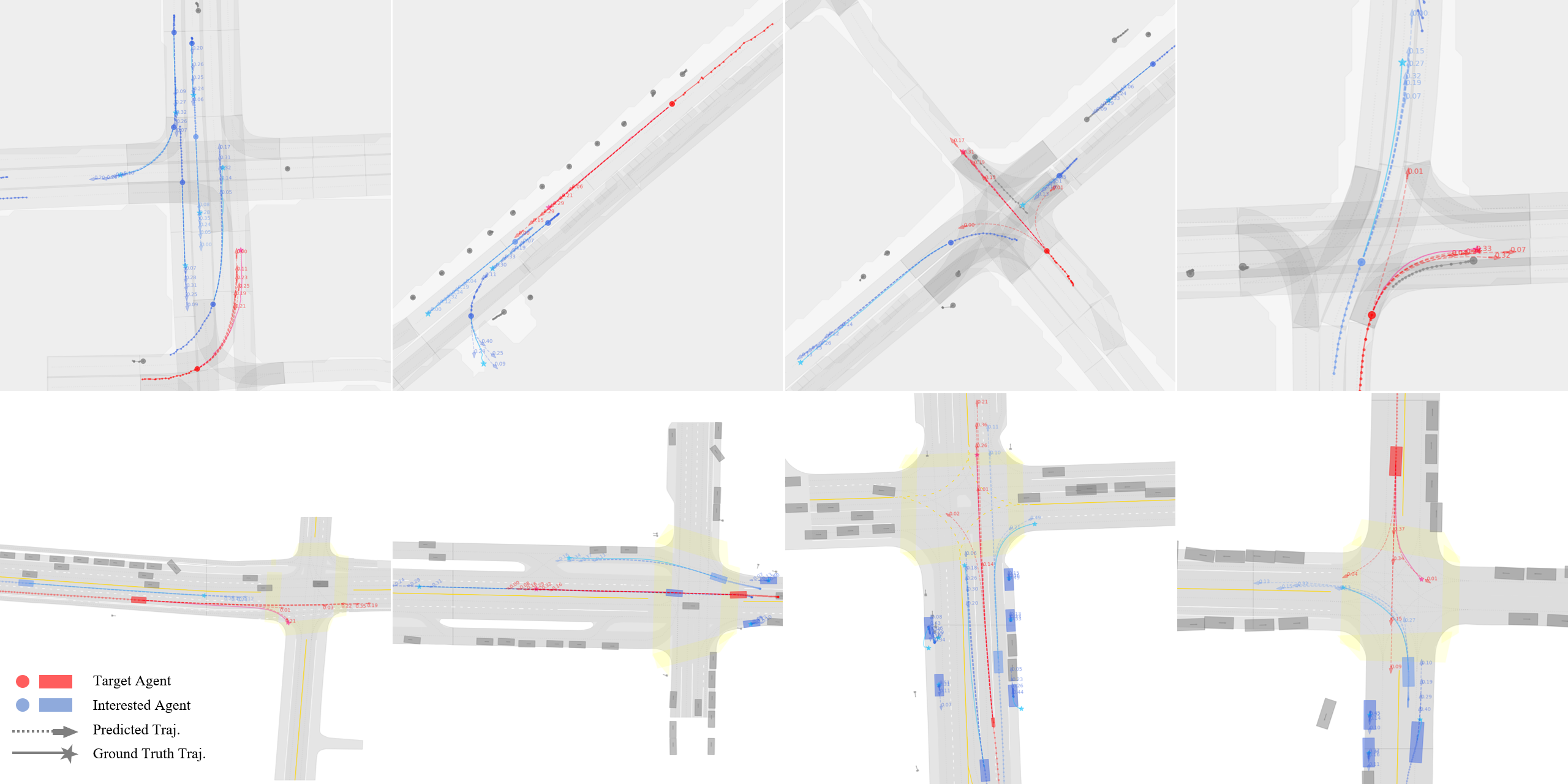}
	\caption{Qualitative results on Argoverse 1 (upper) and Argoverse 2 (lower) motion forecasting datasets. The target agents are shown in red while other interested agents are marked in blue. Note that the future motions of all agents in the scene are generated but we omit the results of ignored agents (grey) for conciseness. The ground truth endpoints are denoted as stars and the predicted trajectories are depicted as dashed curves with the final poses marked as arrows. SIMPL effectively extracts driving context features and generates multiple agent trajectories that adhere to the specific scene constraints in complex scenarios. (Please zoom in for details.)}\label{fig:qual_res}
\vspace{-0.6cm}
\end{figure*}

The qualitative results on both Argoverse 1 and 2 datasets are illustrated in Fig.~\ref{fig:qual_res}. Our SIMPL is able to anticipate realistic, reasonable, and accurate multimodal future trajectories for multiple agents in the scene simultaneously. We also demonstrate the qualitative results of real-time consecutive trajectory prediction on the Argoverse tracking dataset based on models trained on the motion forecasting dataset without fine-tuning (zero-shot transfer). The snapshots are depicted in Fig.~\ref{fig:cover}, and for detailed results please refer to the attached supplementary video.

\subsection{Ablation Study}
\subsubsection{On feature fusion module}

We first investigate the design of the proposed SFT layer. As shown in Tab.~\ref{tab:abl_sft}, with the growth of embedding size and number of SFT layers ($\mathcal{M}1$$\rightarrow$$\mathcal{M}3$), SIMPL achieves better performance in all metrics. However, given the embedding size of 128, increasing the number of SFT layers from 4 to 6 ($\mathcal{M}4$$\rightarrow$$\mathcal{M}5$) marginally improves the prediction accuracy at the cost of incorporating $22\%$ more parameters, which is less preferred in real-time applications. We also find that updating the relative positional embedding (RPE) using the context array in each layer can boost the overall performance significantly ($\mathcal{M}3$$\rightarrow$$\mathcal{M}4$). We surmise that this is due to the fact that updating the RPE involves incorporating node features into edge features, which helps to learn the relationship between different semantic instances.

\begin{table}[!t]
    \centering
    \caption{Ablative study of the feature fusion module design on the Argoverse 1 validation split.\label{tab:abl_sft}}
    \setlength{\tabcolsep}{1.0mm}
    \begin{tabular}{c|ccc|ccc}
    \toprule
    Model          & Emb. Size & $\#$ Layers  & RPE Upd.   & minFDE$_{6}$    &  MR$_{6}$       & b-minFDE$_{6}$   \\
    \midrule
    $\mathcal{M}1$ & 64        & 2            & \xmark     & 1.237           &  12.8           & 1.848          \\
    $\mathcal{M}2$ & 64        & 4            & \xmark     & 1.037           &  9.5            & 1.658          \\
    $\mathcal{M}3$ & 128       & 4            & \xmark     & 0.993           &  9.0            & 1.607          \\
    $\mathcal{M}4$ & 128       & 4            & \cmark     & 0.947           &  \textbf{8.1}   & 1.559          \\
    $\mathcal{M}5$ & 128       & 6            & \cmark     & \textbf{0.944}  &  8.4            & \textbf{1.558} \\
    \bottomrule
\end{tabular}
\vspace{-0.4cm}
\end{table}

\begin{table}[!t]
    \centering
    \caption{Ablative study of the trajectory parameterization methods and the yaw angle loss on the Argoverse 2 validation set.}\label{tab:abl_traj}
    \setlength{\tabcolsep}{1mm}
    \begin{tabular}{l|c|cccc}
    \toprule
    Parameterization  & Yaw loss  & minADE$_{6}$    & minFDE$_{6}$    & minAYE$_{6}$    & minFYE$_{6}$     \\
    \midrule
    Raw coords        & \xmark    & \textbf{0.780}  & \textbf{1.452}  & 0.134           &  0.151           \\
    Polynomial        & \xmark    & 0.861           & 1.738           & 0.146           &  0.278           \\
    B\'ezier curve	  & \xmark    & \textbf{0.780}  & 1.457           & 0.137           &  0.297           \\
    B\'ezier curve	  & \cmark    & 0.783           & \textbf{1.452}  & \textbf{0.055}  &  \textbf{0.076}  \\
    \bottomrule
\end{tabular}
\end{table}

\subsubsection{On trajectory parameterization}\label{sec:abl_param}
\begin{figure}[t]
	\centering
	\includegraphics[width=0.48\textwidth]{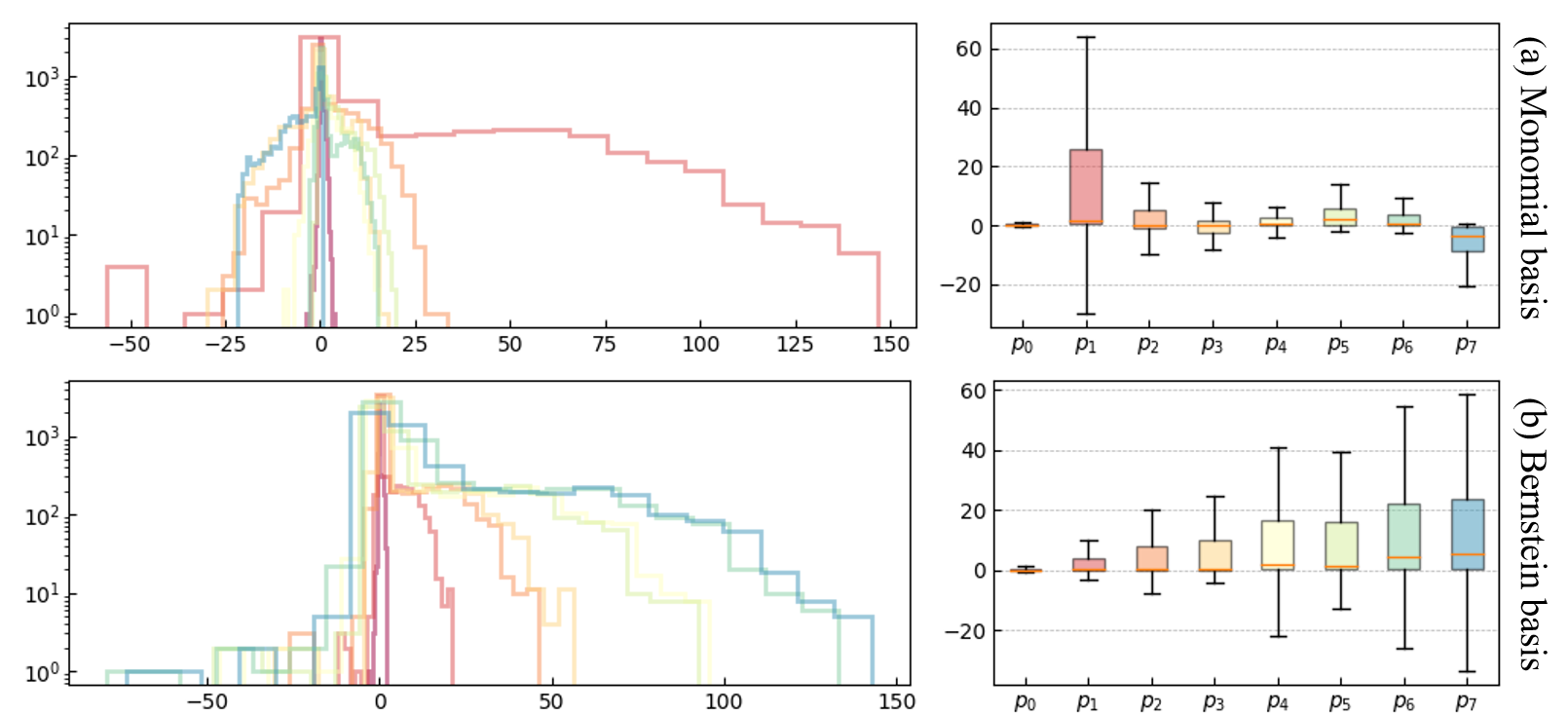}
	\caption{Distribution of the predicted $x$-axis coefficients of monomial basis polynomials (a) and Bernstein basis polynomials (b). The left column shows the histogram of the coefficient distribution while the right column shows the corresponding boxplot. The distributions of coefficients of different orders are indicated by different colors. We can find that the $1$st-order coefficient of the monomial basis polynomial has a much wider span than others, while the coefficients (i.e., control points) of the B\'ezier curves are more regular due to their specific spatial meaning.}\label{fig:coeff_dist}
\end{figure}

We further show the influence of different trajectory parameterization methods in Tab.~\ref{tab:abl_traj}. Similar to the conclusion described in~\cite{vara2022multipath++}, the monomial basis polynomial representation brings a significant performance drop compared with raw coordinates. In contrast, our B\'ezier curve-based method achieves the same level of results in the displacement-related metrics. We blame the performance drop on the numerical imbalance of the monomial basis, while the coefficients of B\'ezier curves are control points with specific spatial meanings, without significant difference in difficulty from directly regressing coordinates. We compare the predicted coefficient distributions for different parameterization methods (see Fig.~\ref{fig:coeff_dist}), and it shows the distribution of B\'ezier curve is more regular than the monomial basis, potentially making this task easier.

\subsubsection{On auxiliary loss functions}
Leveraging continuous representations also makes it more convenient for us to access higher-order physical quantities without violating physical constraints. Therefore, we can naturally introduce loss functions for quantities such as heading angles without any modification to the network architecture. To evaluate the yaw loss introduced in~\ref{sec:training}, we introduce the minimum average yaw error (minAYE$_k$) and minimum final yaw error (minFYE$_k$), which directly calculate the absolute angular difference in radians. From Tab.~\ref{tab:abl_traj}, we can clearly find that yaw loss substantially improves the accuracy of yaw angles, which is highly favorable to real-world applications.

\subsection{Extensibility}
As described above, our SIMPL follows the simplest possible network architecture design, leaving spaces for further extensions. To demonstrate this merit, we attach an additional simple trajectory decoder following the idea of iterative proposal refinement, which is widely adopted in recent state-of-the-art approaches~\cite{choi2022r, shi2022motion, zhou2023query}. We apply no modification to the vanilla SIMPL and regard the predicted multimodal trajectories as initial proposals. For simplicity, here we only refine the predicted trajectories of the target agent. Similar to~\cite{choi2022r, shi2022motion}, for each proposal trajectory, we first re-encode it into a feature vector. Then, we collect the nearby instance features of each proposal within a certain range, followed by a feature fusion module based on the standard Transformer decoder, attending the local context to the corresponding proposal feature. After the refinement, the proposal features are sent to another simple MLP-based decoder to get the final predicted trajectories and their probability scores. The training process is identical to the vanilla SIMPL, namely, employing the WTA strategy with the best-predicted proposal as the positive trajectory. We denote the enhanced model as SIMPL-R, and the results are shown in Tab.~\ref{tab:ext_refine}. With such a simple plug-and-play post-refinement module, SIMPL-R achieves better overall performance, indicating the compact architecture makes it scalable and promising to be used as a backbone for a variety of different tasks. We also note that SIMPL can be smoothly integrated with other recent techniques such as self-supervised learning~\cite{bhattacharyya2023ssl, cheng2023forecast}. We leave it as another future work.

\begin{table}[tb]
	\centering
	\caption{Quantitative results of extensibility experiment on the Argoverse 1 validation and test split.\label{tab:ext_refine}}
	\setlength{\tabcolsep}{1.75mm}
	\begin{tabular}{c|c|cccc}
	\toprule
    Split               & Method    & minADE$_{6}$     & minFDE$_{6}$   & MR$_{6}$      & b-minFDE$_{6}$  \\
	\midrule
    \multirow{2}*{Val}  & SIMPL     & 0.658            & 0.947          & \textbf{8.1}  & 1.559           \\
                        & SIMPL-R   & \textbf{0.651}   & \textbf{0.946} & 8.2           & \textbf{1.542}  \\
        \midrule
    \multirow{2}*{Test} & SIMPL     & 0.793            & 1.179          & 12.3          & 1.809           \\
                        & SIMPL-R   & \textbf{0.783}   & \textbf{1.173} & \textbf{12.1} & \textbf{1.781}  \\
  \bottomrule
\end{tabular}
\end{table}

\section{Conclusion}\label{sec:concl}
In this paper, we present a simple and efficient multi-agent motion prediction baseline for autonomous driving. Leveraging the proposed symmetric fusion Transformer, the proposed method achieves efficient global feature fusion and retains robustness against viewpoint shifting. The continuous trajectory parameterization based on Bernstein basis polynomials provides higher compatibility with downstream modules. The experimental results on large-scale public datasets show that SIMPL is more advantageous in terms of model size and inference speed while obtaining the same level of accuracy as other state-of-the-art methods.


\end{document}